\documentclass[preprint,12pt,authoryear]{elsarticle}\setcitestyle{numbers,square,sort&compress,comma}
\usepackage{url}
\usepackage{hyperref}
\usepackage{lineno}
\usepackage{booktabs}
\usepackage{multirow}
\usepackage{pdfcomment}  
\usepackage{amssymb}
\usepackage{amsmath}

\journal{Arxiv}

\begin{document}

\begin{frontmatter}

\title{GANORM: Lifespan Normative Modeling of EEG Network Topology based on Multinational Cross-Spectra} 

\author[1]{Shiang Hu\corref{cor1}}
\ead{shu@ahu.edu.cn}

\author[1]{Xiaolong Huang}

\author[1]{Yifan Hu}
\author[1]{Xue Xiang}
\author[1]{Xiaoliang Sheng}
\author[2]{Debin Zhou}
\author[3,4]{Pedro A. Valdes-Sosa}

\affiliation[1]{organization={Anhui Provincial Key Lab of Multimodal Cognitive Computation, Key Lab of Intelligent Computing and Signal Processing of Ministry of Education, School of Computer Science and Technology, Anhui University}, postcode={230601},city={Hefei},country={China}}
\affiliation[2]{organization={Stony Brook Institue at Anhui University},postcode={230601},city={Hefei},country={China}}
\affiliation[3]{organization={School of Life Science and Technology, University of Electronic Science and Technology of China},
                postcode={611731},city={Chengdu},country={China}}
\affiliation[4]{organization={Department of Neuroinformatics, Cuban Neuroscience Center},postcode={11600},city={Havana},country={Cuba}}

\cortext[cor1]{Corresponding author at: School of Computer Science and Technology, Anhui University, 230601, Hefei, China}

\begin{abstract}
\noindent \textbf{Background and Objective:} Charting the lifespan evolutionary trajectory of brain function serves as the normative standard for preventing mental disorders during brain development and aging. Although numerous MRI studies have mapped the structural connectome for young cohorts, the EEG-based functional connectome is unknown to characterize human lifespan, limiting its practical applications for the early detection of brain dysfunctions at the community level.

\noindent \textbf{Methods:}
This work aimed to undertake normative modeling from the perspective of EEG network topology. Frequency-dependent scalp EEG functional networks were constructed based on EEG cross-spectra aged 5-97 years from 9 countries and network characteristics were quantified. First, GAMLSS were applied to describe the normative curves of the network characteristics in different frequency bands. Subsequently, addressing the limitations of existing regression approaches for whole brain network analysis, this paper proposed an interpretable encoder-decoder framework, Generative Age-dependent brain Network nORmative Model (GANORM). Building upon this framework, we established an age-dependent normative trajectory of the complete brain network for the entire lifespan. Finally, we validated the effectiveness of the norm using EEG datasets from multiple sites.

\noindent \textbf{Results:} After building normative curves of network characteristics, we observed a more pronounced change in the $\alpha$ band, suggesting a significant role of $\alpha$ band in brain development and aging process. Subsequently, we evaluated the effectiveness of GANORM, and the tested  performances of BPNN showed the $R^2$ was 0.796 ± 0.0191, the MAE was 0.081 ± 0.0009, and the RMSE was 0.013 ± 0.0006. Following established lifespan brain network norm, GANORM also exhibited good results upon verification using healthy and disease data from various sites. The deviation scores from the normative mean for the healthy control group (MNCS, HC, HY-AHU, 0.1015-0.1385) were significantly smaller than those of the disease group (BrainLat datasets, 0.1672-0.1869).

\noindent \textbf{Conclusion:} This study employed EEG cross-spectra for normative modeling, which will potentially exert a significant impact on disease diagnosis and individual deviation analysis. The evolutionary curves of normative network characteristics and topological patterns of scalp EEG network across the lifespan offered standardized references from large-scale cohort for individual stratification of disease severity.
\end{abstract}


\end{frontmatter}


\section{Introduction}
Normative modeling is an emerging research topic in population neuroscience, which helped to understand population variations at the individual level, but not at the average level\cite{shan2022mapping}. Like height or weight growth charts in pediatric medicine\cite{bethlehem2022brain}, normative modeling establishes the relationship between neuroimaging-derived characteristics and demographic variables such as age or sex\cite{bozek2023normative,rutherford2023evidence}. Subsequently, individual deviations from the norms are used to prevent, diagnose, and treat mental disorders \cite{porras2021predictive,marquand2019conceptualizing,dimitrova2020heterogeneity,bethlehem2020normative,wolfers2018mapping}. Current normative modeling studies primarily utilize MRI anatomical data, providing an effective tool to capture structural changes, but being low sensitive to brain dynamics. Due to the advantages of non-invasiveness, portability, high temporal resolution, and excellent sensitivity to brain dynamics, electroencephalography (EEG) will exclusively play a vital role in creating the normative functional connectome.

Numerous methods for normative modeling were proposed, with Gaussian process regression (GPR) being one of the most widely used \cite{marquand2019conceptualizing,rutherford2024normative}. Dimitrova et al. established norms for the microstructure of the neonatal brain and found that preterm infants exhibited significant deviations\cite{dimitrova2020heterogeneity}. Zabihi et al. used GPR to estimate longitudinal normative models for cortical thickness and then mapped individual deviations from the typical pattern\cite{zabihi2019dissecting}. Wolfers et al. employed GPR to predict gray matter volumes by age and gender, estimating normal brain changes throughout the adult period\cite{wolfers2020individual}. Gur et al. used linear regression to create age-related cognitive growth charts\cite{gur2014neurocognitive}. Kessler et al. proposed a novel method for predicting attention deficit using brain network growth charts based on polynomial regression
\cite{kessler2016growth}. Erus et al. estimated support vector regression models for brain development \cite{erus2015imaging}. Kia et al. employed a hierarchical Bayesian regression framework within a federated probabilistic setting to perform normative modeling of lifespan, offering the potential to decentralized neuroimaging data\cite{kia2022closing}. Fraza established normative models for neuroimaging-derived features in a large cohort using warped Bayesian linear regression. Recently, Generalized additive models for location, scale and shape(GAMLSS) has been recognized as an effective approach for normative modeling \cite{bozek2023normative}.
Bethlehem et al. have mapped precise brain charts for the grey matter in a large cohort \cite{bethlehem2022brain}. Subsequently, Rutherford et al. conducted validation using warped Bayesian regression on medical data, demonstrating significant clinical value\cite{rutherford2022charting}.

Although numerous studies have conducted normative modeling of neuroimaging features, most have focused on modeling brain structural changes. Functional connectivity (FC) is known to change continuously throughout development and aging, which is useful to examine individual differences and group analysis in pathology\cite{sun2019graph,duan2020topological,nogales2023discriminating,elam2021human}. Some studies have performed normative modeling on FC. Sun et al. used GPR to establish a normative model of FC strength across the lifespan, revealing individual heterogeneity in patients with major depressive\cite{sun2023mapping}. Rutherford et al. used PCNToolkit to establish a normative model of FC and investigated differences between schizophrenia and healthy control groups\cite{rutherford2023evidence}. However, previous regression approaches are not suitable for regression at the whole brain level. They can be applied to univariate attributes, but face difficulties in conducting standardized modeling for the whole brain network.

Recently, functional developmental connectome studies have provided some insight into functional variation using complex network analysis\cite{adebisi2024eeg,diykh2020eeg,bajestani2019diagnosis,zuo2017human,oldham2019development,grayson2017development}. These studies calculated network characteristics(NCs) such as characteristic path length and global efficiency considering integration and segregation\cite{farahani2019application,van2019cross}, and have summarized developmental patterns for certain age groups. During infancy, the brain structure is immature, and adolescence is the fastest stage of development, both characterized by continuous changes in the brain. Gilmore et al. demonstrated that functional networks in the sensorimotor resting state are present at birth, while functional networks of higher order gradually emerge and develop within the first two years of life\cite{gilmore2018imaging}. Khundrakpam et al. found significant changes in topological characteristics in later childhood, particularly a significant decrease in local efficiency and modularity, and a significant increase in global efficiency\cite{khundrakpam2013developmental}. Damoiseaux et al. have demonstrated a strong correlation between age effects and both structural and functional connectivity, with older adults generally exhibiting lower within-network connectivity and higher between-network connectivity\cite{damoiseaux2017effects}.Furthermore, Cao et al. found that modularity decreases linearly, while local efficiency and rich club architecture follow an inverted U-shaped trajectory in the age range of 7-85\cite{cao2014topological}. Deery et al. observed that the brains of older adults have less modularity and higher integration but lower efficiency at rest\cite{deery2023older}.Most of the studies listed above focus on fMRI data, while the potential of EEG used for normative modeling is largely unknown, although EEG is an important neuroimaging modality with high temporal resolution and portability at a lower cost. Especially the early diagnosis of brain disorders urgently requires EEG biomarkers\cite{engemann2020combining}. 

Deep generative models, used as normative models for identifying neurological disorders in the brain, have demonstrated significant results\cite{zhang2024dcnet,tang2024learning,qayyum2023high,chen2022modern,yildiz2022unsupervised}. The encoder-decoder architecture is a widely used neural network architecture in deep learning for sequence-to-sequence tasks, such as natural language processing, computer vision, and speech recognition\cite{badrinarayanan2017segnet,ji2021cnn}. The encoder is responsible for transforming the input raw data into low-dimensional features, while the decoder converts these low-dimensional features back into the original data space. Brain-computer interfaces (BCIs) also embody this architecture, enabling the conversion of brain signals into commands for operating computers\cite{ingolfsson2024brainfusenet,xu2021review}. Furthermore, research has applied the encoder-decoder architecture to the analysis of EEG signals, such as the dynamic encoder-decoder architecture proposed by Arefnezhad et al., which estimates driver drowsiness through EEG\cite{arefnezhad2022driver}. Sayantan et al. proposed multi-modal variational autoencoder (mmVAE) based normative modeling framework that can capture the joint distribution between different modalities and apply it for normative modeling\cite{kumar2023normative}. In this study, we drew inspiration from the encoder-decoder architecture to establish a lifespan whole brain network norm. Considering the unique nature of EEG as medical data analysis, which demands higher interpretability\cite{you2020unsupervised}, we utilized specialized expert knowledge in EEG signal processing and brain network analysis for the encoding part, while leveraging neural networks in the decoding part for target transformation of features.

This study performed normative modeling based on the EEG network on a large sample of resting-state EEG, employing GAMLSS to depict the evolutionary trajectory of NCs across the lifespan. proposed a model of encoder-decoder architecture to establish the evolutionary trajectory of whole brain networks across the entire lifespan. The findings revealed that the NCs exhibited different nonlinear trajectories over the lifespan, with varying connectivity strengths across different ages. This study described the normative evolutionary topological patterns across the lifespan, providing further insight into the process of brain development and aging, and serving as a promising model for quantifying individual stratification. In this paper, the main contributions are summarized as follows:
\begin{itemize}
\item[$\bullet$] \textbf{Novel application domain}: First EEG Network topology-based normative model complementing MRI-centric approaches, enabling direct assessment of electrophysiological dynamics. 
\item[$\bullet$] \textbf{Methodological advancement}: First implementation of Generalized Additive Models for Location, Scale, and Shape (GAMLSS) to derive lifespan trajectories for EEG network topology.
\item[$\bullet$] \textbf{Translational tool development}: We design an interpretable normative modelling tool (GANORM) based on encoder-decoder architecture for brain networks at whole brain level, leveraging our expertise in EEG cross-spectral signal processing. Based on GANORM and large-scale multinational EEG Cross-spectral dataset, we established for the first time a whole brain network norm across the lifespan. 
\item[$\bullet$]  \textbf{Clinical validation}: Cross-site validation using heterogeneous healthy and clinical datasets, demonstrating utility in individualized deviation analysis for neuropsychiatric disorders.

\end{itemize}

The remainders are structured as follows: Section \ref{section2} introduces data and data analysis methods. Section \ref{section3} presents our data analysis results, along with cross-site validation results. The discussion follows in Sections \ref{section4}, and ends up with conclusion in Section \ref{section5}.

\section{\bf Material and Methods}\label{section2}
\subsection{\bf MNCS dataset}
The MNCS dataset originates from the global EEG Norm Project of the Global Brain Consortium (GBC) (\url{https://3design.github.io/GlobalBrainConsortium.org/project-norms.html}), which employed a decentralized data sharing strategy. Each site is not required to share raw EEG, but only to share cross-spectra (CS) processed by the unified scripts, along with anonymized subject information that includes age and other demographic details. The MNCS dataset has an almost uniform gender distribution. The CS is a frequency-domain function between two signals, which is used to express the amplitude components and phase relationships of the common frequency components contained within these two signals\cite{zhang2023applied}.

\begin{figure}[h]
\centering
\includegraphics[width=3.4 in]{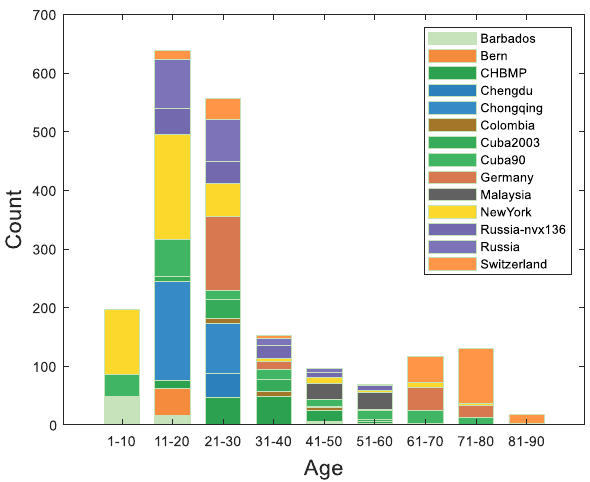}
\caption{\textbf{Age distribution of MNCS dataset}.}
\label{fig1}
\end{figure}

\begin{figure*}[htp]
\centering
\includegraphics[width=5 in]{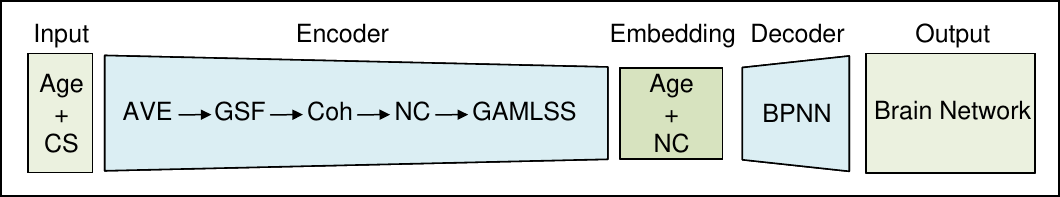}
\caption{\textbf{Generative Age-dependent brain Network nORmative Model (GANORM)}. The input is EEG cross-spectra with age. Based on neuroscientific methods, the encoder performs a series of processing on EEG, including average reference, global factor correction, construction of coherence-based scalp EEG network, calculation of NCs, and GAMLSS modeling of NCs. Embedding is age and normative NCs (50\% quantile after GAMLSS modeling). The decoder is a five-layer BPNN that inputs age and normative NCs and outputs normative brain networks at each age.}
\label{fig2}
\end{figure*}

Fig\ref{fig1} is the age distribution of the MNCS dataset, which includes 1966 subjects from 9 countries across 14 sites. Age values cover almost the entire lifespan. The raw EEG recordings were from the 10/20 international electrode placement system with 19 electrodes:Fp1, Fp2, F3, F4, C3, C4, P3, P4, O1, O2, F7, F8, T3/T7, T4/T8, T5/P7, T6/P8, Fz, Cz, Pz. The frequency range is 1.17-19.14Hz with 0.39Hz interval. The age ranges 5-97 yrs., skewed toward younger subjects. The research workflow is illustrated in Fig\ref{fig2}. 

\subsection{\bf Preprocessing}
The data preprocessing included two steps referred to as \textit{AVE} and \textit{GSF} in the encoder: average referencing and global-scale factor correction\cite{li2022harmonized}. The raw EEG recording that the CS originated from may be set to inconsistent references, hindering the standardization in later normative modeling. The \textit{AVE} transformed all CS matrices $\bf{S}_i(\omega)$ from the inconsistent online recording reference or the offline re-references to the common virtual average reference\cite{hu2019statistics,yao2019reference,hu2018reference,hu2018unified}:

\begin{equation}
\label{eq1}
\tilde{\bf{S}}_i(\omega)=\bf{HS}_i(\omega)\bf{H}^T
\end{equation}

\noindent where $\bf{H}=\bf{I}_{N_c}-\bf{1}_{N_c}{\bf{1}_{N_c}}^T/N_c$ is the average reference transformation matrix, $\omega$ is frequency. Although two EEG recordings may show a similar appearance, they may differ significantly in overall amplitude in the actual EEG recording, resulted from different amplifiers, recording conditions, and devices. This interference was solved by dividing the CS by the stochastic global scale factor described by Hernandez et al.\cite{li2022harmonized}. To analyze the relationship between signals in different frequency bands and brain activity states, we divided the CS into four frequency bands: $\delta$(1.17-3.12Hz), $\theta$(3.51-7.81Hz), $\alpha$(8.20-12.1Hz) and $\beta$(12.5-19.14Hz)\cite{hu2024xi}.

\subsection{\bf Scalp EEG network}
This section refers to the \textit{Coh} of the encoder. The CS matrices were stored into a CS tensor by stacking up frequencies. Since original EEG data are unavailable, FC can only be estimated using measures that can be derived from CS\cite{nolte2020mathematical}. Coherence, as normalized CS amplitudes, is a widely applied measure to estimate FC\cite{li2023coherence}. The coherence matrix reflects the multivariate cross-correlation between multiple signals in the frequency domain after taking the Fourier transformation. Using the coherence matrix as the FC adjacency matrix, the weighted undirected network was created with the electrodes serving as nodes, with pairwise coherence values representing the connectivity strength between electrodes. The coherence per frequency $\omega$ was calculated as follows:
\begin{equation}
    \label{eq2}
    Coh_{xy}(\omega)=\frac{{\left| P_{xy}(\omega) \right|}^2}{P_{xx}(\omega) \cdot P_{yy}(\omega)}
\end{equation}
Here, $P_{xy}(\omega)$ represents the CS density of signals x and y, $P_{xx}(\omega)$ and $P_{yy}(\omega)$ represent the power spectral density of signals x and y. Through calculations, we obtain the coherence FC matrices for all subjects at each frequency, each matrix having dimensions of $N_c*N_c*N_f$, where $N_c$ denotes the number of electrodes and $N_f$ represents the number of frequency points.

\subsection{\bf Age-dependent trajectories of Network characteristics}\label{NC norm}
This section describes the \textit{NC} and \textit{GAMLSS} of the encoder. The graph theory based complex network analysis, such as NCs increasingly plays a significant role in the prediction and treatment of various mental disorders. NCs encompasses integration metrics such as characteristic path length (CPL) and global efficiency (GE), segregation metrics such as clustering coefficient (CC), local efficiency (LE) and modularity(M), and centrality metrics such as betweenness centrality(BC) and participation coefficient(PC), representing the stability and transmission efficiency of the network. 

Specifically, CPL refers to the average length of all the shortest paths between all pairs of nodes as a feature of the global network. It signifies the efficiency of information transmission across the network, with smaller values indicating higher efficiency. GE is inversely proportional to the average shortest path length and measures the communication efficiency of a network. CC is the ratio of the actual number of connected edges to the most possible number of connected edges, used to quantify the neighbor strength of connected nodes in a network as a local network feature. LE measures the local information transmission ability of a network and reflects the network's ability to defend against random attacks to a certain extent. M is the degree to which the network may be subdivided into such clearly delineated and non-overlapping groups. BC measures the role of a node as a \textit{bridge} in a network, and nodes with high betweenness centrality are key nodes for the flow of information and resources. Nodes with high PC may facilitate global modular integration. The calculation of all NCs used the scripts from the Brain Connectivity toolbox\cite{rubinov2010complex}. 

The general additive models for location, scale and shape (GAMLSS) provides a comprehensive description of the data distribution by constructing flexible regression models for multiple distribution parameters of the response variable and allows modeling nonlinear relationships within the data. GAMLSS was used to construct the evolutionary trajectories of NCs across four frequency bands throughout the lifespan. We use log-transformed age values as the independent variable and established the evolutionary trajectories of seven NCs using cubic smoothing splines and Box-Cox-t distribution family. Additionally, we set percentile lines at 5\%, 25\%, 50\%, 75\%, and 95\% to better understand the distribution of evolutionary trajectories across different percentiles.

\subsection{\bf Normative brain network based on BPNN regression}
The NC-based evolutionary trajectories can provide valuable insight into the variation in brain function with age, but these NCs hardly capture the complexity and dynamics of brain networks. To gain a more comprehensive understanding of the evolutionary patterns of scalp EEG network topologies, we depict the evolutionary trajectories of brain networks at the whole brain level. A BPNN regression model was designed, with a network architecture consisting of an input layer, three hidden layers (h1, h2, h3), and an output layer. Note that the number of neurons in each layer is determined on the basis of the embedding selection, which is a novelty of our study. The overall concept of the BPNN involves utilizing the fused features of functional EEG NCs such as CPL, GE, CC, LE, M, BC, and PC at a specific frequency band or a frequency point corresponding to a subject’s chronological age, as input for the model. The upper triangular elements of the FC matrix at the same specific frequency band or frequency point for the same subject are then used as the output of the model. The compact FC can be reconstructed from the predicted upper triangular elements of the FC matrix learned from BPNN.

\subsubsection{\bf Embedding Selection}
\begin{figure*}[htp]
\centering
\includegraphics[width=5 in]{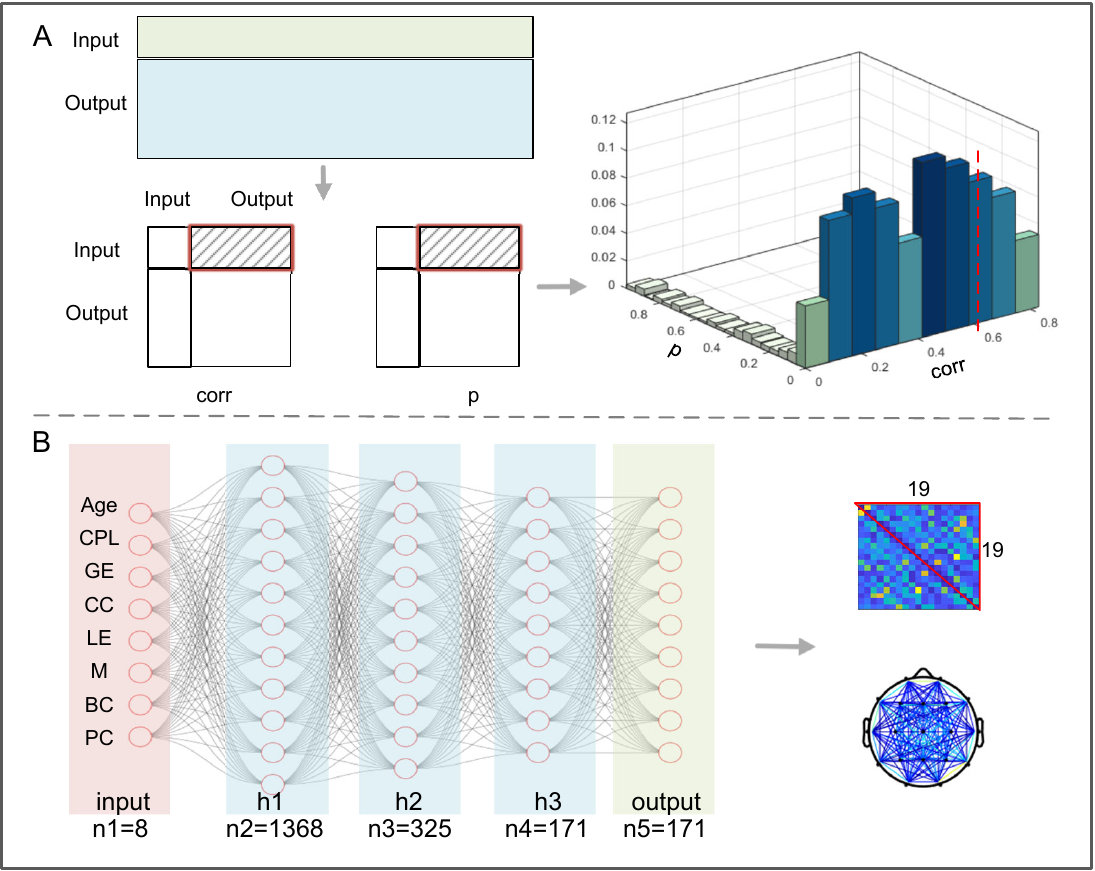}
\caption{\textbf{Design of the backpropagation neural network (BPNN)}. (A)Input-output Correlation analysis. Among the 8 input features $\times$ 171 output pairwise FC strengths, 325 show a correlation \textgreater 0.6 and p-values \textless 
 0.05. (B) BPNN diagram. A five-layer BP neural network with an input layer of 8 neurons, three hidden layers with 1368, 325, and 171 neurons respectively, and an output layer of 171 neurons.}
\label{fig3}
\end{figure*}

This section pertains to the construction of the decoder. The goal of embedding selection was to analyze the association between the model inputs and outputs. The inputs were chronological age and the NCs that included CPL, GL, CC, LE, M, BC, and PC, with size $ R^{8*1} $, and the outputs corresponded to the upper triangular elements in the FC matrix, which consists of 171 pairwise FC weights. Thus, the Pearson correlation coefficient between any one from the 8 model inputs and any one from the 171 pairwise FC weights were calculated across the 1945 subjects. The p-values between each input of the model using age or 7 NCs against each output of the model using pairwise FC weight were also reported. Statistically, we find that 325 features have a correlation \textgreater 0.6 with p-values \textless 0.05 as shown in Fig\ref{fig3}A, indicating that the BPNN can effectively extract the relationship between features and upper triangular elements using 325 parameters.

\subsubsection{\bf BPNN module}
Our objective is to predict the entire adjacency matrix of the FC using the chronological age of a subject and the age-dependent normative mean NCs. That is, only the chronological age of the subject is required, and the other inputs are the normative mean NCs picked from the NCs trajectories in Section\ref{NC norm}. Thus, the BPNN is configured with an input layer with 8 neurons and an output layer with 171 neurons, the latter of which is equal to the size of entries in the upper triangular of the FC matrix. The network includes three hidden layers with the ReLU activation function. The first hidden layer has 1368 neurons (8*171), indicating that each input element will correspond to a parameter to learn the relationship with each output element. The second hidden layer has 325 neurons, which is the number of strong correlations we found between the 8 input features and the output elements. The third hidden layer has 171 neurons, representing the upper triangular elements of the brain network that the BPNN aims to predict.

\subsubsection{ \bf Normative brain network prediction} 
The BPNN utilizes the Adam optimization algorithm to adaptively adjust the learning rate for each parameter. Five-fold cross-validation was employed to evaluate the model's generalization capability. We subsequently evaluate the model's performance using R-squared ($R^2$), Mean Absolute Error (MAE), and Root Mean Squared Error (RMSE) as follows:
\begin{equation}
\label{eq3}
R^2=1-\frac{\underset{i}{\sum}(\hat{y}_i-y_i)^2}{\underset{i}{\sum}(\overline{y}_i-y_i)^2}
\end{equation}
\begin{equation}
\label{eq4}
MAE=\frac{1}{m}\sum_{i=1}^{m}\left| y_i-\hat{y}_i \right|
\end{equation}
\begin{equation}
\label{eq5}
RMSE=\sqrt{\frac{1}{m}\sum_{i=1}^{m}(y_i-\hat{y}_i)^2}
\end{equation}
where $y_i$ and $\hat{y}_i$ represent the actual value and the predicted value in formulas \ref{eq3} - \ref{eq5}, $\overline{y}_i$ in \ref{eq3} means the mean value, and $m$ in \ref{eq4} and \ref{eq5} represents the number of statistical observations. The parameters of the BPNN regression model were learned after training. To chart evolutionary patterns of the brain network across the lifespan, the age-dependent normative brain network can be constructed by taking any chronological age and the age-dependent normative mean NCs that we learned from the GAMLSS model as input. The normative mean NCs are just the 50\% values of NCs learned by GAMLSS, reflecting the inherent NCs that the normative brain network should present.

\subsection{\bf Application of GANORM}
Using the GANORM model, we established normative brain networks for each age group. To validate the effectiveness of GANORM, we collected 58 healthy young individual EEG data (age range 19-25 years) under resting conditions from Anhui University (HY-AHU). All informed consents were signed from the participants. Then we obtained access to the BrainLat dataset(age range 21-89), which includes 530 patients with neurodegenerative diseases such as Alzheimer’s disease (AD), behavioral variant frontotemporal dementia (bvFTD), multiple sclerosis (MS), Parkinson’s disease (PD), and 250 healthy controls (HC)\cite{prado2023brainlat}. We processed all data uniformly and computed their brain networks and NCs. Subsequently, we calculated the deviations between subjects and the established normative brain networks from two perspectives: mean functional connection strength (MFCS) and NCs, and conducted statistical analyses. The deviation in NCs refers to the difference between the NCs of the subject and those of the normative network. The formula for computing the deviations of MFCS for each subject is as follows:
\begin{equation}
\label{eq6}
Deviation_{MFCS(i)}=\frac{\displaystyle\sum_{m=1}^{N_c}\sum_{n=1}^{N_c}(\bf{FC}(m,n)_{subject(i)}-\bf{FC}(m,n)_{norm})}{N_c*N_c}
\end{equation}
where $FC_{subject(i)}$ represents the functional connectivity matrix of the subject(i), $FC_{norm}$ represents the generated normative functional connectivity matrix for the corresponding age, and Nc represents the number of channels, i.e., the dimension of the matrices, m and n represent the rows and columns of FC. 

\section{\bf Results}\label{section3}
\subsection{\bf Trajectory of network characteristics}
After applying the GAMLSS to NCs, we obtained the nonlinear evolutionary trajectories of NCs across four frequency bands($\delta$, $\theta$, $\alpha$, $\beta$) across the lifespan. Then, we summarized and analyzed the NCs from three perspectives: functional integration, functional segregation, and centrality.

\begin{figure*}[htp]
\centering
\includegraphics[width=5 in]{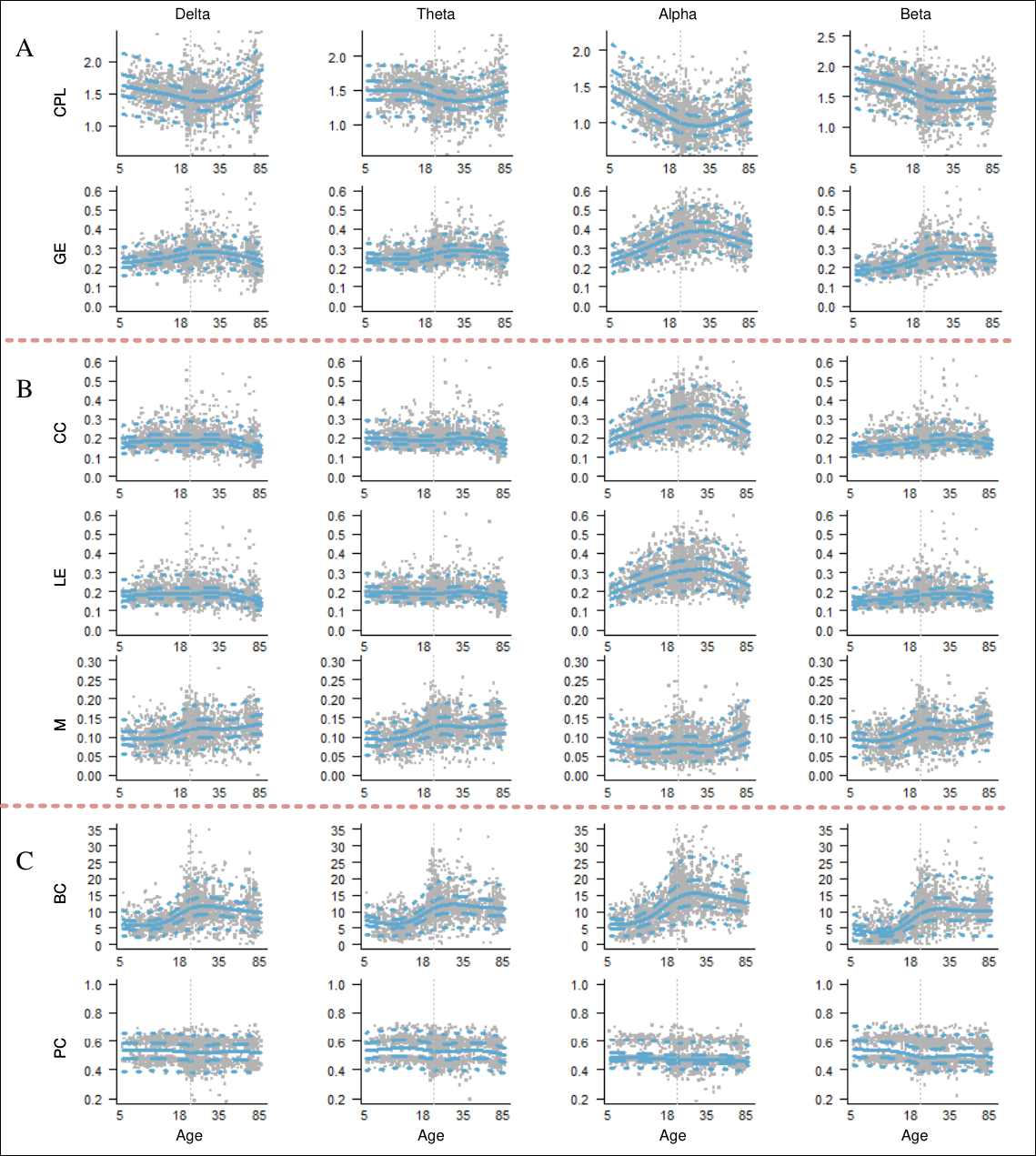}
\caption{\textbf{Trajectory of network characteristics}. A, B, and C represent functional integration, functional segregation and centrality, respectively. Different rows correspond to different network characteristics, while different columns correspond to different frequency bands. CPL-characteristic path length, GE-global efficiency, CC-clustering coefficient, LE-local efficiency, M-modularity coefficient, BC-betweenness centrality, PC-participation coefficient vs. age in log scale as the independent variable. All fits from bottom to up, percentiles at 5\%, 25\%, 50\%, 75\%, and 95\% display evident nonlinear patterns.}
\label{fig4}
\end{figure*}

In terms of functional integration (Fig \ref{fig4}(A)), the CPL decreases and then increases with age, reaching a minimum around the age of thirty, while GE increases and then decreases with age, peaking around the age of thirty. These trends are most pronounced in the alpha band. These findings collectively suggest that the brain's information integration capacity is relatively weak during adolescence, strengthens with age, becomes mature and stable in adulthood, and declines in old age.

In terms of functional segregation (Fig \ref{fig4}(B)), the CC and LE initially increase slightly with age in the delta and beta bands, plateauing in adulthood and slightly decreasing in old age. The alpha band exhibits a similar trend, but with a notably faster rate of change. In contrast, the theta band remains relatively stable until around the age of fifty, after which there is a slight decreasing trend.These phenomena suggest that during adolescence, the brain's local connectivity and information processing capabilities are relatively weak, which gradually mature and stabilize during adulthood, with a slight decline observed in old age. Additionally, we can observe that the alpha band plays a more prominent role in this aspect. The M in the delta, theta, and beta bands exhibits a consistently nonlinear increase with age, while in the alpha band, it initially decreases and then increases during adolescence, and similarly, it shows a slight decrease followed by an increase in adulthood, which can be considered an approximate steady state. Overall, the modular structure of the network shows a slightly upward trend across the lifespan.

In terms of centrality analysis (Fig \ref{fig4}(C)), BC exhibits similar trends across the four frequency bands, with a slight increase during adolescence, a rapid increase peaking around the age of thirty, and a subsequent slight decline. The PC remains stable throughout the lifespan across all four frequency bands. In our centrality analysis, we only consider the overall efficiency of the network at the average level, thereby overlooking the functional roles of individual nodes or brain regions at different stages. However, this provides a foundation for our subsequent neural network normative modeling.

Overall, various functions of the brain network, including the efficiency and speed of information transfer and the specialized processing of information across brain regions, are not fully developed in childhood, with all functions being relatively weak. During adolescence, these functions experience a rapid growth rate, reaching a peak around the age of thirty, and then remain stable throughout adulthood. In old age, there is a slight decline. The alpha band plays a significant role in the brain network's functions, regulating various aspects of brain network functionality.

\subsection{\bf Evaluation of the BPNN}
To validate the rationality of the number of neurons in each layer of the BPNN, we varied the number of neurons in the hidden layers (h1,h2) to mimic the ablation experiment and conducted five-fold cross-validation. \textit{Model} refers to the model we designed. \textit{Model/2} indicates that the number of neurons in the h1 and h2 layers is halved from the original 1368 and 325, respectively. \textit{Model*2} means the number of neurons in each layer is doubled. \textit{10-10} denotes that both the h1 and h2 layers have 10 neurons. During the training phase, our model exhibited the best performance ($R^2$: 0.814±0.0052, MAE: 0.079±0.0002, RMSE: 0.011±0.0007). In the testing phase, it also demonstrated superior performance ($R^2$: 0.796±0.0191, MAE: 0.081±0.0009, RMSE: 0.013±0.0006), $R^2$ was only 0.005 lower than \textit{Model/2}. The results demonstrate that our model achieves the best performance, confirming the rationality of the set number of neurons in the hidden layers. Notly that our study is the first to conduct interpretable normative modeling of brain networks at the whole brain level, constituting an interpretable image generation approach. In contrast, existing image generation techniques primarily rely on deep learning, whose learning processes act as black boxes, lacking transparency and interpretability. Consequently, we are without an appropriate baseline for comparative effectiveness analysis.

\begin{table}[htp]
    \centering
    \caption{Model Evaluation}
    \label{tab1}
    \begin{tabular}{ccccc}
    \toprule
        ~ & ~ & $R^2$ & MAE & RMSE \\ 
        \midrule
        \multirow{2}*{10-10} & 
            train & 0.804±0.0063 &  0.081±0.0005 & 0.012±0.0023 \\
        ~ & test & 0.797±0.0160 & 0.082±0.0017 & 0.013±0.0046 \\
        \midrule
        
        \multirow{2}*{Model/4} & 
            train & 0.813±0.0026 &  0.080±0.0005 & 0.012±0.0040 \\
        ~ & test & 0.792±0.0328 & 0.081±0.0015 & 0.014±0.0029 \\
        \midrule
        
        \multirow{2}*{Model/2} & 
            train & 0.810±0.0116 &  0.080±0.0012 & 0.015±0.0032 \\
        ~ & test & \bf{0.801±0.0183} & 0.081±0.0019 & 0.015±0.0023 \\
        \midrule
        
        \multirow{2}*{\textbf{Model}} & 
            train & \bf{0.814±0.0052} & \bf{0.079±0.0002} & \bf{0.011±0.0007}\\
        ~ & test & 0.796±0.0191 & \bf{0.081±0.0009} & \bf{0.013±0.0006} \\
        \midrule
        
        \multirow{2}*{Model*2} & 
            train & 0.806±0.0094 &  0.080±0.0014 & 0.013±0.0039 \\
        ~ & test & 0.794±0.0165 & 0.081±0.0018 & 0.014±0.0033 \\
    \bottomrule
    \end{tabular}
\end{table}

\subsection{\bf Evolutionary patterns of brain network}
\begin{figure*}[htp]
\centering
\includegraphics[width=5.1 in]{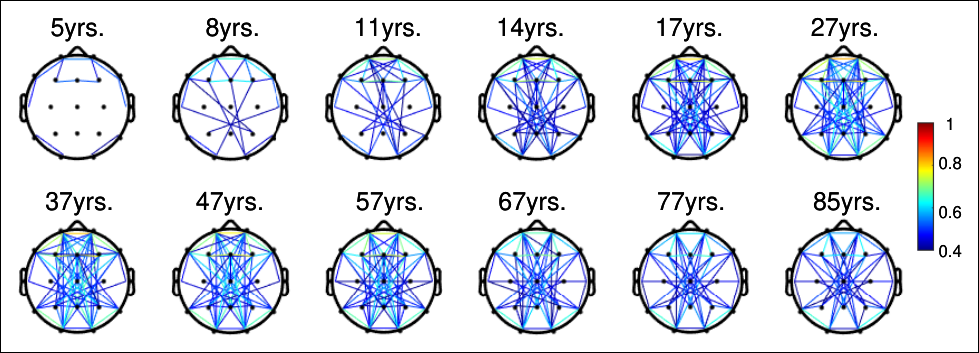}
\caption{\textbf{Lifespan evolutionary patterns of scalp EEG network topology in $\alpha$ band}. Normative network topologies from ages 5 to 85 are depicted, with brain networks for selected ages shown in the figure. The age interval is 3 years before age 17 and 10 years thereafter, with a network connection threshold set to 0.4.}
\label{fig5}
\end{figure*}

 We obtained the evolutionary patterns of the brain network across the lifespan (Fig\ref{fig5}) after inputting the age and the NCs at the 50th percentile by GAMLSS modeling into the trained BPNN. We observed that during childhood, the brain network had fewer connections with weaker strengths. In adolescence, the number of connections gradually increased, and their strengths became stronger. By adulthood, the number and strength of connections reached their peak and remained stable. In old age, the connections gradually weakened, yet the network structure was still able to maintain a stable state.

\begin{figure*}[h]
\centering
\includegraphics[width=5.5 in]{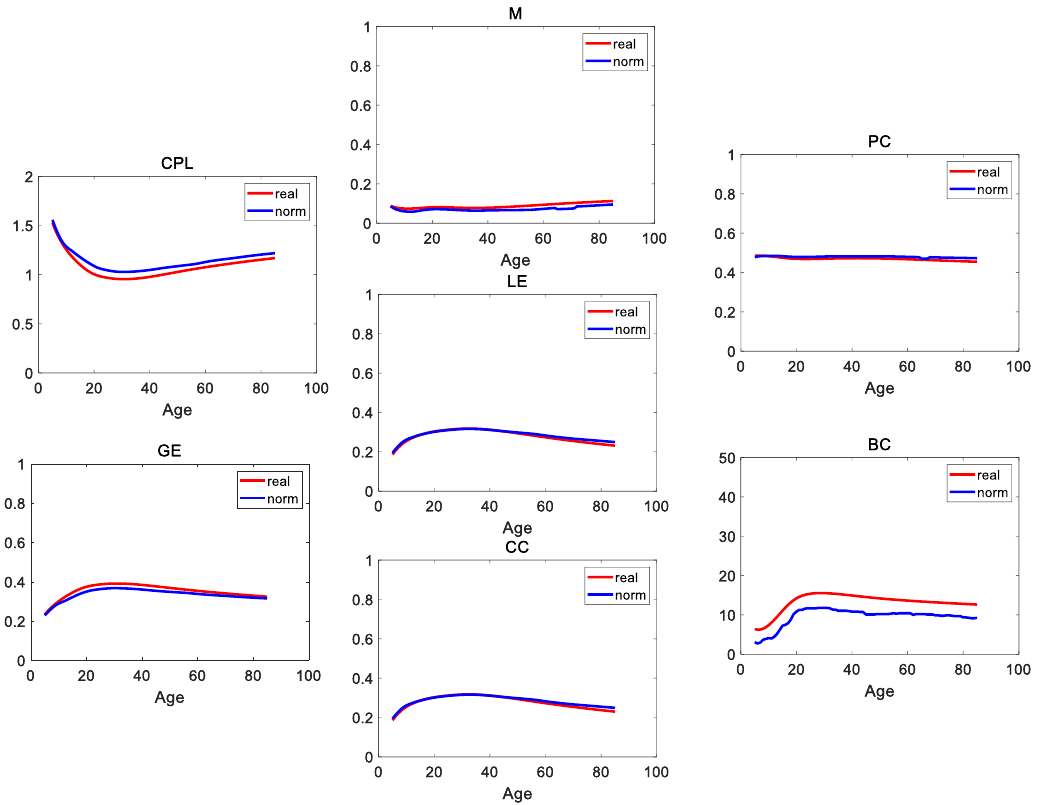}
\caption{\textbf{Comparison of generated normative NCs and actual brain NCs}. It illustrates the comparison of seven NCs between the generated norms and the actual network, with the actual  depicted in red and the normative in blue.}
\label{fig6}
\end{figure*}

After obtaining the brain network norms, we calculated the NCs of the norm and compared them with the NCs inputted into the BPNN. The results indicate that the NCs of the norms closely match those at the 50th percentile, demonstrating that the norms we constructed are in line with reality (Fig\ref{fig6}). The increase of CPL and decrease of BC in the generated network, respectively, are attributed to the thresholding process applied to the brain network during computation. The increase in CPL is due to the thresholding that results in the neglect of finer details and connections within the network, causing some paths between nodes to become disconnected, thereby inflating the CPL. The decrease in BC is a consequence of the thresholding that renders the connections between network nodes relatively sparse, diminishing the \textit{bridge} role of nodes and consequently lowering their BC.

\subsection{\bf Application of GANORM on independent dataset}
\begin{figure*}[htp]
\centering
\includegraphics[width=5 in]{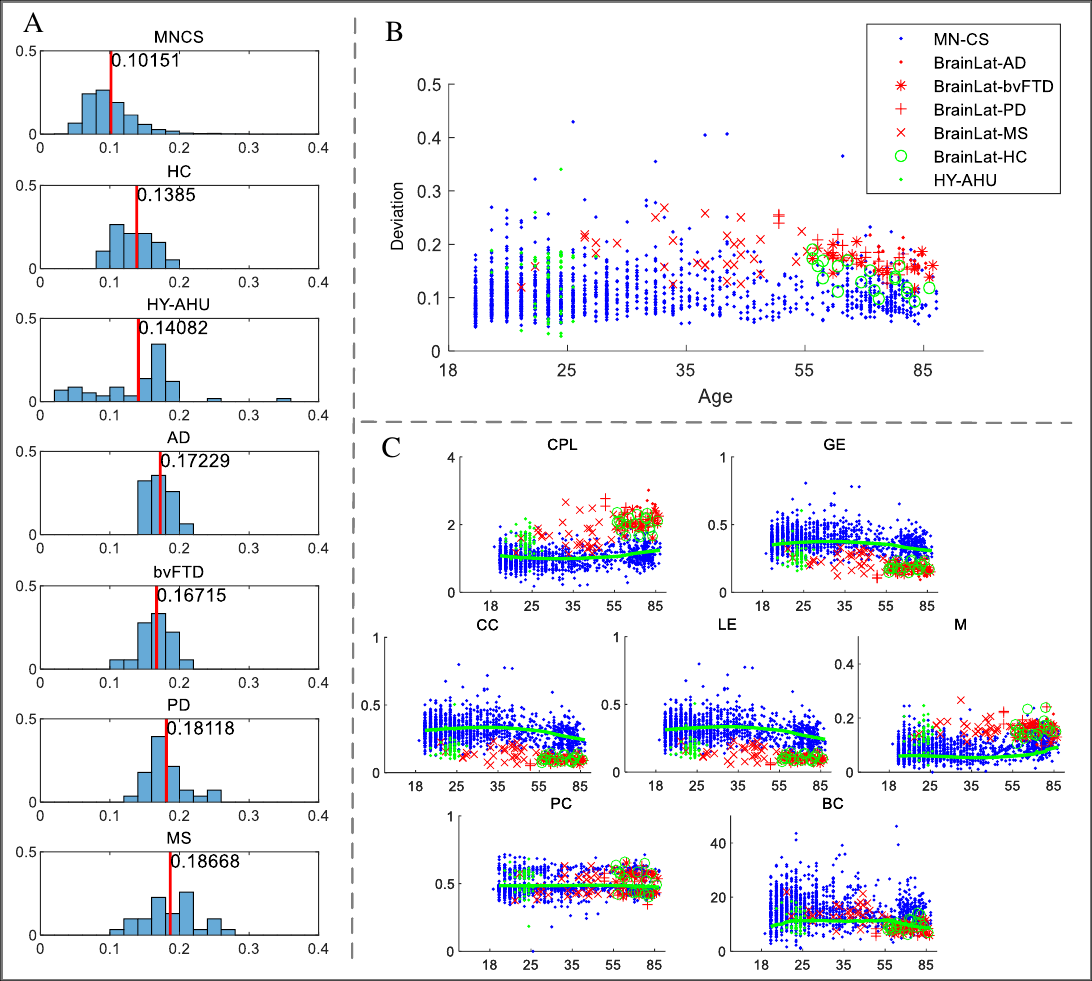}
\caption{\textbf{Cross-site deviation from GANORM}. (A) Average absolute deviations for three groups of healthy data (MN-CS, HC, HY-AHU) and four groups of disease data (AD, bvFTD, PD, MS). The horizontal axis represents deviation values, while the vertical axis represents probability density. (B) MFCS deviations. Blue dots denote MN-CS healthy data, green circled dots represent other healthy data, and red dots are labeled as disease data. (C) Distribution of NCs. The green curve represents the nomative NCs. The vertical axis represents NCs values. }
\label{fig7}
\end{figure*}

We calculated the deviation and conducted statistical analysis for the HY-AHU and BrainLat datasets. Through visualization, we observed that in terms of MFCS, the deviation of the healthy group is significantly smaller than that of the disease group (Fig\ref{fig7}A). Additionally, in terms of NCs, the values of the healthy group are closer to the normative NCs (Fig\ref{fig7}B). Statistical analysis has confirmed these findings. In terms of MFCS, the data from MN-CS (0.1015), HY-AHU (0.1408), and BrainLat-HC (0.1385) exhibit lower deviations, whereas the deviations in the data for AD (0.1723), bvFTD (0.1672), PD (0.1812), and MS (0.1867) from BrainLat significantly higher than those in the healthy control group. In the context of deviations in disease data, the deviations associated with bvFTD are relatively small, whereas the deviations for MS are the largest. After statistical analysis of NCs, it was also found that the NCs of the healthy group were closer to the normative values. The results indicated that our model exhibited good performance.

\section{\bf Discussion}\label{section4}
In this study, we initially constructed a Generative Age-dependent brain Network nORmative Model(GANORM) based on the encoder-decoder architecture and established the evolution trajectory of brain networks across the lifespan using this model. We then validated the normative model's effectiveness through datasets from other sites. Our research focuses on establishing the evolution trajectory-normative model of brain networks across the lifespan for a cross-sectional large sample of healthy control population. Through GANORM, we observed that adolescents exhibit weaker network connectivity and information transmission capabilities, yet their developmental speed is faster. In adults, the connection strength is robust, with strong and stable information transmission capabilities. In contrast, elderly individuals show a certain decline compared to adults. Through validation with cross-site datasets, we found that the generative brain network model performs well, with lower deviations in healthy data and higher deviations in disease data. This provides crucial reference significance for diagnosis and treatment related to brain function in the psychiatry field.

In numerous research endeavors exploring brain aging, cognitive functions, and the field of neuroscience, MRI data are predominantly utilized due to their high spatial resolution, which allows for the revelation of structural details within the brain\cite{gilmore2018imaging}. In this study, we utilized large-scale EEG data to construct brain evolution trajectories based on the following reasons. Firstly, our research objective is to construct the evolution trajectories of brain functional networks, particularly focusing on the electrical activity changes related to cognitive functions, where EEG demonstrates certain advantages\cite{kaccar2016design}. Secondly, we aim to construct evolution trajectories across the lifespan. Collecting data from elderly individuals poses certain challenges, and validating model results necessitates the collection of data from various psychiatric disorders. EEG acquisition is relatively simpler and more suitable for such populations. Lastly, EEG equipment is relatively inexpensive and can be easily used in various environments, which is crucial for conducting large-scale research. In an environment where MRI data are predominantly used in most studies, this study's utilization of EEG data provides a vital perspective for the investigation of brain evolution trajectories.

The GANORM model is based on an encoder-decoder architecture, yet it differs from the traditional sense of such architectures. Traditional encoder-decoders are implemented through recurrent neural networks (RNN) or their variants (such as LSTM and GRU)\cite{ye2019understanding,asadi2020encoder}. However, the \textit{encoder} of our model is accomplished through specialized feature extraction and selection within the field of EEG signal processing. EEG, as a signal in the medical domain, is crucial for an accurate diagnosis and treatment. While deep learning excels in feature extraction, its \textit{black box} operational characteristic makes the specific logic behind predictions elusive\cite{ieracitano2022novel,nogales2023discriminating}. In the medical field, interpretability is vital for clinical decision-making\cite{mortaga2021towards}, and thus this limitation of deep learning can pose problems. To avoid this scenario, we chose to utilize professional knowledge in the field of EEG signals for feature extraction and selection\cite{hu2018reference}. This method not only extracts features highly relevant to the task but also provides neurophysiological explanations behind these features, thereby enhancing the interpretability of the results. This is particularly important for clinicians, as they need to formulate and adjust treatment plans based on these explanations. In summary, we chose to use feature extraction and selection methods from the field of EEG signals during the encoding phase to meet the high demand for interpretability in medical data analysis.

The use of longitudinal versus cross-sectional data has been a major concern. Longitudinal data allows for direct observation of individual changes over time and provides a more accurate reflection of such changes. However, acquiring longitudinal data necessitates prolonged tracking and multiple scans, while numerous external factors (such as lifestyle changes and onset of diseases) can influence the trajectories. This study constructed brain aging charts using cross-sectional data, which offers the advantage of collecting a vast amount of data across different age groups within a short period, providing a broad perspective to observe general trends in brain aging and rendering our findings more representative and universal. Cross-sectional data does have certain limitations, including an inability to directly observe individual changes over time, which may underestimate the actual changes in brain aging\cite{di2023mapping}. In addition, there may be significant variations in the trajectories of brain aging between different individuals, and cross-sectional studies struggle to accurately capture these differences. Although this study relied primarily on cross-sectional data to construct brain aging charts, we also recognize the potential value of longitudinal data. Future research can further explore the integration of these two types of data to more fully delineate the trajectories of brain aging\cite{chen2021neuroimaging}. Additionally, it is worth investigating a framework for the utilization of cross-sectional data, which can be adapted for the assessment of longitudinal data\cite{buvckova2025using}.

\section{\bf Conclusion}\label{section5}
We used EEG data for normative modeling, offering an alternative crucial perspective to MRI-based normative modeling. This paper proposes a lifespan brain network evolution trajectory construction model, GANORM, based on an encoder-decoder architecture. During the encoding phase, we extracted network attribute features using specialized EEG signal analysis theories and feature extraction methods within the field of EEG. Specifically, we employed the GAMLSS to establish normative evolution trajectories for lifespan NCs, uncovering significant variability patterns in the alpha frequency band. Subsequently, we generated a brain network for the NCs within the alpha band using BPNN regression, thereby establishing a normative evolution trajectory for the lifespan brain network. We then collected two datasets (HY-AHU and BrainLat) for cross-site validation. The results demonstrated that our model exhibited good generalization ability, and the constructed normalized model achieved high accuracy. This study provides an important perspective for understanding the developmental trajectory of brain networks across the entire lifespan and offers significant insights for the diagnosis and prediction of mental disorders. The MATLAB packages and tutorial are freely available at \url{https://github.com/ShiangHu/GANORM}.

\section*{Declaration of generative AI and AI-assisted technologies in the writing process}
During the preparation of this work the authors used Grammarly in order to grammar checking. After using this tool, the authors reviewed and edited the content as needed and take full responsibility for the content of the publication.

\section*{CRediT authorship contribution statement}
\textbf{Shiang Hu}: Writing, editing, Supervision, Funding acquisition, Resources, Conceptualization, Methodology, Investigation, Formal analysis.
\textbf{Xiaolong Huang}: Writing, editing, Software, Visualization, Methodology, Investigation, Formal analysis, Validation.
\textbf{Yifan Hu}: Investigation, Validation.
\textbf{Xue Xiang}: Investigation, Validation.
\textbf{Xiaoliang Sheng}: Investigation, Validation.
\textbf{Debin Zhou}: Investigation.
\textbf{Pedro A. Valdes-Sosa}: Dataset Collection.

\section*{Declaration of competing interest}
The authors declare that they have no known competing financial interests or personal relationships that could have appeared to influence the work reported in this paper.

\section*{Acknowledgments}
This work was supported by the National Natural Science Foundation of China [Grant No. 62101003]. 

 \bibliographystyle{unsrt} 
 \biboptions{sort&compress}
 \bibliography{ref}

\end{document}